\documentclass[letterpaper, 10 pt, conference]{ieeeconf}  

\IEEEoverridecommandlockouts                              

\usepackage{graphics} 
\usepackage{graphicx} 
\usepackage{amsmath} 
\usepackage[noend]{algpseudocode}
\usepackage{algorithmicx,algorithm}
\usepackage{cite}
\usepackage{multirow}
\usepackage{subfigure}
\usepackage{hyperref}
\usepackage{color,xcolor}
\usepackage{array}

\title{\LARGE \bf
ROI-based Robotic Grasp Detection for Object Overlapping Scenes
}

\author{Hanbo Zhang, Xuguang Lan, Site Bai, Xinwen Zhou, Zhiqiang Tian and Nanning Zheng
\thanks{Hanbo Zhang and Xuguang Lan are with the Institute of Artificial Intelligence and Robotics, the National Engineering Laboratory for Visual Information Processing and Applications, School of Electronic and Information Engineering,
        Xi'an Jiaotong University, No.28 Xianning Road, Xi'an, Shaanxi, China.
        {\tt\small zhanghanbo163@stu.xjtu.edu.cn, xglan@mail.xjtu.edu.cn}}%
}

\begin{document}

\maketitle
\thispagestyle{empty}
\pagestyle{empty}

\begin{abstract}
Grasp detection considiering the affiliations between grasps and their owner in object overlapping scenes is a necessary and challenging task for the practical use of the robotic grasping approach. In this paper, a robotic grasp detection algorithm named ROI-GD is proposed to provide a feasible solution to this problem based on Region of Interest (ROI), which is the region proposal for objects. ROI-GD uses features from ROIs to detect grasps instead of the whole scene. It has two stages: the first stage is  to provide ROIs in the input image and the second-stage is the grasp detector based on ROI features. We also contribute a multi-object grasp dataset, which is much larger than Cornell Grasp Dataset, by labeling Visual Manipulation Relationship Dataset. Experimental results demonstrate that ROI-GD performs much better in object overlapping scenes and at the meantime, remains comparable with state-of-the-art grasp detection algorithms on Cornell Grasp Dataset and Jacquard Dataset. Robotic experiments demonstrate that ROI-GD can help robots grasp the target in single-object and multi-object scenes with the overall success rates of 92.5\% and 83.8\% respectively.
\end{abstract}

\section{Introduction}

In pursuit of a broader coverage on synergic functions in our daily life, it is essential for service robot and industrial robot to be competent in grasping a specified object in complex scenes containing multiple objects. Very often, robots are required to grasp the relevant target, specifically demanded, in specified scenarios. Therefore, an efficient algorithm that accomplishes the grasp for a specified target in complex scenes is the premise for extensive application of robot in our everyday lives. The {\bf main challenges} to solve this problem include: 1) grasp detection with the affiliation between grasps and their owner in multi-object scenes; 2) guarantee of other objects' safety during grasping; 3) robust motion planning and control. In this paper, we focus on the first challenge. In detail, we divide this challenge into the following two parts:
\begin{itemize}
\item {\bf How to locate the grasps in a stack of objects}: when objects are in disorganized heaps, there are overlaps and occlusions between objects, which makes grasp detection extremely difficult.
\item {\bf How to know which object each grasp affiliates to}: after grasps are detected, it is difficult to match the detected grasps with corresponding objects due to the overlaps in multi-object stacking scenes. 
\end{itemize}
In this paper, the owner of a grasp in multi-object scenes is defined as the object to which the grasp is affiliated.

 \begin{figure}[t] 
 \center{\includegraphics[scale=0.08]{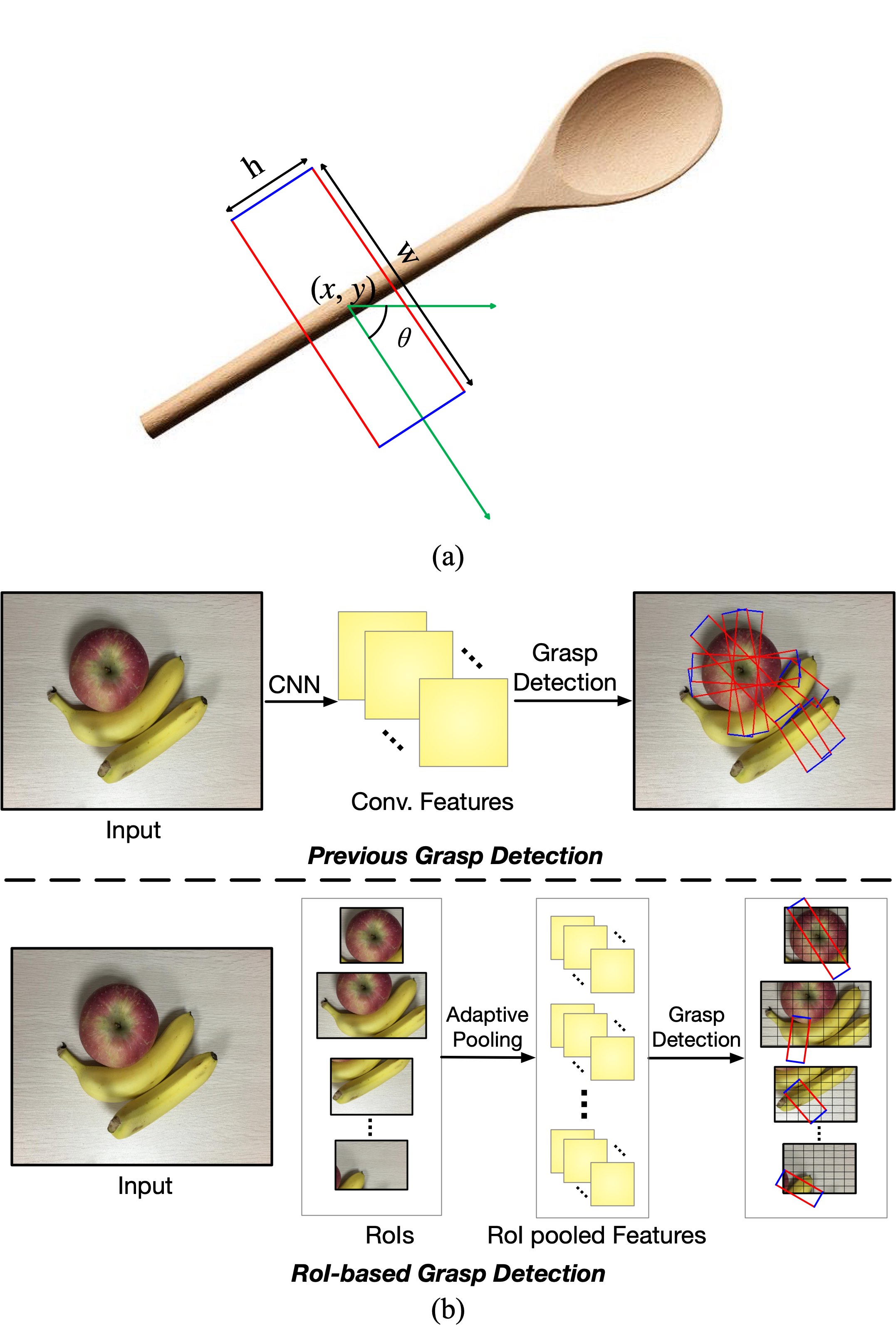}}        
 \caption{(a) Grasp representation in this paper. Each grasp includes 5 dimensions: $(x,y,w,h,\theta)$. $(x,y)$ is the coordinate of the center point. $(w,h)$ is the width and height of the grasp rectangle. $\theta$ is the rotation angle with respect to the horizontal axis. (b) Grasp detection on ROIs compared with previous grasp detection algorithms. Grasp detection on ROIs takes two steps: First, use ROI pooling to get a batch of ROI pooled features with same size $W\times H$ ($e.g.$ $7\times7$), and all ROIs are also divided into $W\times H$ grid cells. Second, use grasp detector to detect grasps on ROIs. {\bf The main difference}: the ROI-based grasp detection detects grasps using ROI features instead of features of the whole image.}
 \label{intro}
 \end{figure}

Recent works\cite{redmongrasp, resnetgrasp, chu2018real, zhou2018grasp} focus on grasp detection in single object scenes. However, in practical applications of robotic grasping, the scenes faced by robots often contain multiple objects. Some works\cite{handeyegrasp,gualtieri2016high,mahler2017grasp,mahler2018dexnet3,zeng2018robotic} try to solve the problem of grasping in object cluttered scenes, which is also known as the ``bin picking" problem. However, they are unsuitable for situations where the affiliations between grasps and their owner are required. For example, if requested for an apple, these algorithms will be invalid due to the unawareness of which grasp affiliates to the apple. Another alternative is to train the network for grasp detection and object recognition of the most exposed object one by one\cite{guo2016object}. However, the one-by-one detection process limits the efficiency when it is used in multi-object scenes.
 
For humans, grasp detection is object-based in most cases. In detail, grasps should be detected on objects instead of the whole input image which is currently used in existing grasp detection algorithms. Naturally, the simplest way is to break the problem down into two sub-problems: object detection and grasp detection. However, following this method, grasps belonging to different objects may confuse the detector and lead to failures when there are occlusions and overlaps between objects, especially when objects are stacked together.

In this paper, a new approach of robotic grasp detection based on Region of Interest (ROI) named ROI-GD is proposed to solve the problem of grasp detection in object overlapping scenes. ROI refers to the region proposal of the image that is most likely to contain an object. ROI-GD includes two stages: the first stage is to provide ROIs in the input image and the second-stage is the grasp detector based on ROI features. In other words, we use features of the ROI to detect grasps belonging to this area as shown in Fig. \ref{intro}(b), which we call ROI-based robotic grasp detection.
 
For ROI-GD to be trained on multi-object scenes, we need to construct a multi-object grasping dataset. For this reason, Visual Manipulation Relationship Dataset (VMRD), which originally contains 17688 object instances and their category, location and manipulation relationship\cite{zhang2018visual}, are expanded with grasps following the five dimension grasp representation proposed in \cite{jiang2011grasp}. Besides the location of each grasp, the owner of each grasp is also labeled using object index. More than $10^5$ grasps are labeled manually. The five dimension grasp representation of grasp is shown in Fig. \ref{intro}(a). 
 
In order to construct the affiliation between specified objects and their grasps in multi-object scenes, we design a network based on ROI-GD to simultaneously detect grasps and objects and test it on VMRD. Results show that ROI-GD achieves a much better performance for grasp detection in multi-object scenes. Moreover, we train and test our algorithm on Cornell Grasp Dataset and Jacquard Dataset. Results show that ROI-GD performs comparably with state-of-the-art grasp detection algorithms in single-object scenes. 


In summary, contributions of this paper include:

\begin{itemize}
\item A new ROI-based robotic grasp detection approach called ROI-GD is proposed to detect the location of robotic grasps on ROI feature maps. It is comparable with state-of-the-art grasp detection algorithms on single-object grasp tasks while performs much better in object overlapping scenes.
\item A new network is proposed using ROI-GD to simultaneously detect targets and grasps with the affiliation between them in object overlapping scenes.
\item We contribute a new multi-object grasp dataset by expanding Visual Manipulation Relationship Dataset\cite{zhang2018visual} with grasps and the affiliation between grasps and their owner. More than $10^5$ grasps are labeled manually following the grasp representation proposed in \cite{jiang2011grasp}.
\end{itemize}

\section{Related Work}

Robotic grasping has been explored for a long time\cite{bicchi2000robotic, miller2003automatic, graspit, svmgrasp, lu2018planning}. Recently, deep learning provides the possibility of detecting grasps directly from RGB or RGB-D images\cite{lenzgrasp, redmongrasp, resnetgrasp, pinto2016grasp, guo2016grasp, guo2017hybrid, chu2018real, zhou2018grasp,depierre2018jacquard} with its powerful ability of feature extraction\cite{dl}. They treat grasps as a specific kind of object and transfer object detection algorithms to grasp detection, and achieve state-of-the-art performance on single-object grasp datasets such as Cornell Grasp Dataset, CMU Grasp Dataset and Jacquard Dataset. Chu et al.\cite{chu2018real} and Zhou et al.\cite{zhou2018grasp} also explore the performance of deep grasp detection network trained on Cornell Grasp Dataset in multi-object scenes of the real world with no overlap between objects. These algorithms are trained on single-object datasets, which means they have limited performance on object overlapping scenes.

To apply robotic grasping in practice, some works deal with grasping in object cluttered scenes\cite{handeyegrasp, gualtieri2016high, mahler2018dexnet3, mahler2017grasp, zeng2018robotic}. Sergey et al.\cite{handeyegrasp} establish a CNN model for predicting the grasping success rate of the gripper and realize the visual-based continuous servoing control for grasping combined with the CEM algorithm to search in the robot motion space. Gualtieri et al.\cite{gualtieri2016high} search proper grasp pose in clutter through generating a large set of candidates and classifying them into ``good" or ``bad". In \cite{mahler2017grasp}, Mahler et al. model grasping in clutter (also called ``bin picking") as a POMDP and use reinforcement learning to solve the problem. Later, they explore the grasping based on vacuum end effectors in multi-object scenes\cite{mahler2018dexnet3}. In \cite{zeng2018robotic}, authors propose a grasp-first-then-recognize work-flow for robotic grasping and it achieves high performance in the stowing task. However, though they are powerful to find out proper grasps in clutter and good at accomplishing tasks like ``cleaning table" and ``bin picking", it is difficult to apply such kind of algorithms in situations where the affiliations between grasps and their owner are needed. In \cite{guo2016object}, the proposed algorithm recognizes the most exposed object while detects the best grasp simultaneously one by one. When applied to multi-object scenes, it is of low efficiency due to the one-by-one detection process.

Therefore, it is still challenging for grasp detection with the affiliations between grasps and their owner in object overlapping scenes. In our work, we propose an ROI-based robotic grasp detection algorithm. The algorithm we present performs much better in multi-object scenes while remains comparable with state-of-the-art grasp detection algorithms in single-object scenes. 

 \begin{figure*}[!t] 
 \center{\includegraphics[width=0.95\textwidth]{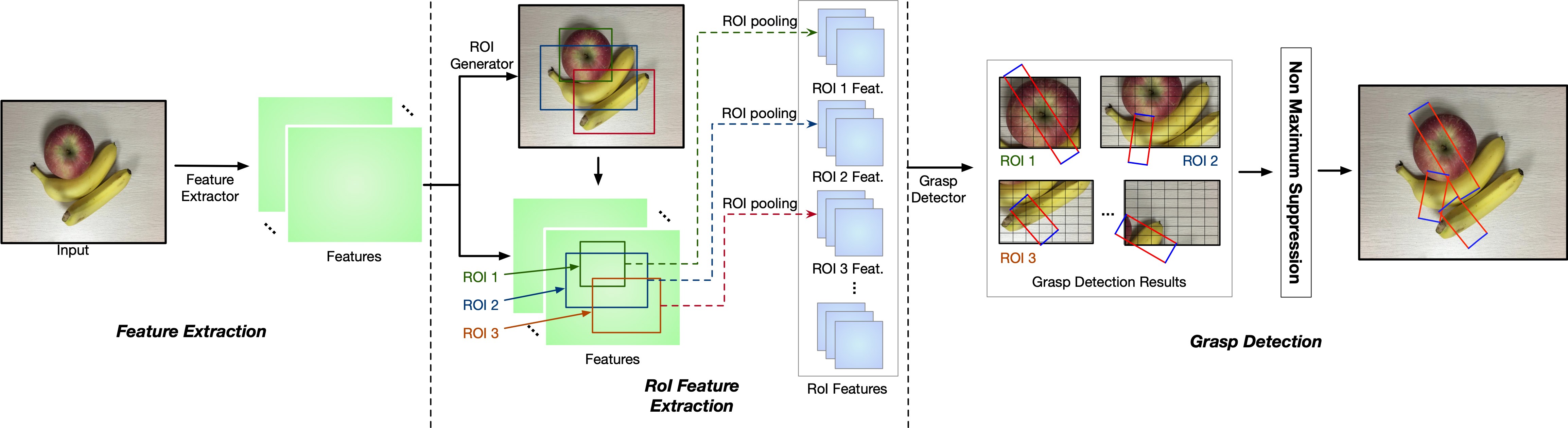}}        
 \caption{Architecture of ROI-GD. All the grasps are detected on ROIs instead of the whole image.}
 \label{arch}
 \end{figure*}
 
\section{Problem Formulation}

The robotic grasp detection problem can be formulated to find proper grasp configurations in the input RGB image. For images including only one object (like images in Cornell Grasp Dataset or Jacquard Dataset), the best configuration $g$ of the object can be formulated using a 5-dimension vector following Lenz et al.\cite{lenzgrasp}:
\begin{equation}
g=(x,y,w,h,\theta)
\end{equation}
where $(x,y)$ is the center of the grasp rectangle, $w$ and $h$ represent the opening size and width of the end effector and $\theta$ denotes the orientation relative to the horizontal axis. Using such a 5-dimension representation of grasp configuration makes grasp detection analogous to object detection. For images including multiple objects (like images in VMRD), a proper grasp should be discovered for each object. Therefore, there is a set of proper grasp configurations $G$, where each element represents a proper grasp configuration for one object in the scene:
\begin{equation}
G=\{g_1,g_2,...,g_n\}
\end{equation}
where $n$ is the number of objects in the scene.

In this paper, we focus on robotic grasp detection in not only single-object but also multi-object scenes.

\section{Proposed Approach}
 
\subsection{Network Architecture}
 
The architecture of ROI-GD is shown in Fig. \ref{arch}. Though it is possible to implement our algorithm without Convolutional Neural Network (CNN) structures, we still adopt CNN because of its strong feature extraction ability. Our network takes RGB images as input, using CNN feature extractor ($e.g. $ ResNet\cite{resnet} or VGG16\cite{vgg}) to extract deep features. Region Proposal Network (RPN), which is in fact three $3\times3$ convolutional layers, provides object bounding box proposals called Regions of Interest (ROIs) for grasp detector (it is denoted by the ``ROI generator" in Fig. \ref{arch}). Afterwards, the ROI pooling layer, as an adaptive pooling layer, pools all features cropped by ROIs into the same size $W \times H$. Finally, a mini-batch of ROI features are fed into grasp detector to detect grasps of every ROI. Note that during testing, our network supports images with different scales, which is different from all previous grasp detection methods.

In this paper, ResNet-101 is used as the backbone. We use $\{C_1,C_2,C_3,C_4,C_5\}$ to denote the features output by each stage's last residual block of the network. ROI features are cropped from C4 and then pooled to $W\times H$. The grasp detector is cascaded after the ROI pooling layer including 3 residual blocks, the grasp rectangle regressor and the classifier. Each ROI is divided into $W\times H$ grid cells. In each grid cell, a set of oriented anchor boxes, which are in fact oriented rectangles used as the priors for grasp regression, are predefined. We use $k$ to represent the anchor number of each grid cell. As shown in Fig. \ref{intro}(a), each grasp is represented by a 5-dimension vector $(x,y,w,h,\theta)$. Therefore, in each grid cell, the grasp rectangle regressor outputs $5\times k$ offsets with respect to the anchors for location of grasp rectangles. Each 5-d vector $(t_x,t_y,t_w,t_h,t_{\theta})$ is associated with an anchor $(x_a,y_a,w_a,h_a,\theta_a)$, and the predicted grasp rectangle $(\hat{x},\hat{y},\hat{w},\hat{h},\hat{\theta})$ is computed using Eq. \ref{encode}:
\begin{equation}
\begin{split}
t_{\hat{x}}=(\hat{x}-&x_{a})/w_{a},
t_{\hat{y}}=(\hat{y}-y_{a})/h_{a},
t_{\hat{w}}=log(\hat{w}/w_a)\\
t_{\hat{h}}&=log(\hat{h}/h_a),
t_{\hat{\theta}}=(\hat{\theta}-\theta_{a})/(90/k)\end{split}
\label{encode}
\end{equation}
The grasp classifier provides $2\times k$ confidence scores, indicating the probabilities of the $k$ anchors to be graspable and ungraspable respectively. For each ROI, the grasp detector will predict $W\times H\times k$ grasp candidates. Finally, by non-maximum suppression (NMS), legal ROIs and the best grasp candidate belonging to them are preserved.

\subsection{Grasp Detection on Regions of Interest}

In ROI-based robotic grasp detection, grasps are detected on ROIs instead of the whole image in order to distinguish grasps belonging to different regions.

ROIs are region proposals of objects\cite{fasterrcnn}. Through ROI pooling, the convolutional features cropped by ROIs are pooled into the same size $W\times H$. In ROI-GD, the network is designed to detect grasps belonging to a specific ROI excluded from grasps on other unspecified ones. 

 
To get training data of an ROI for grasp detection, we first match each ROI with one ground truth bounding box by overlapping area as follow:
\begin{itemize}
\item Keep the ground truth object bounding boxes with Intersection over Union (IoU) larger than 0.5 as matching candidates.
\item Choose the one that has the largest IoU as the matching result.
\end{itemize}
If an ROI does not match any ground truth, it will not be used in training of grasp detection. Then ground truth grasp rectangles belonging to the matched ground truth object will be set as the training data of the ROI. The grasps belonging to other ROIs will be ignored by this ROI even if they are in its area. This encourages the grasp detector to output only  grasps belonging to the specific ROI instead of all potential grasps. 
 
\subsection{Affiliation Construction between Grasps and Objects}

 \begin{figure}[t] 
 \center{\includegraphics[width=0.45\textwidth]{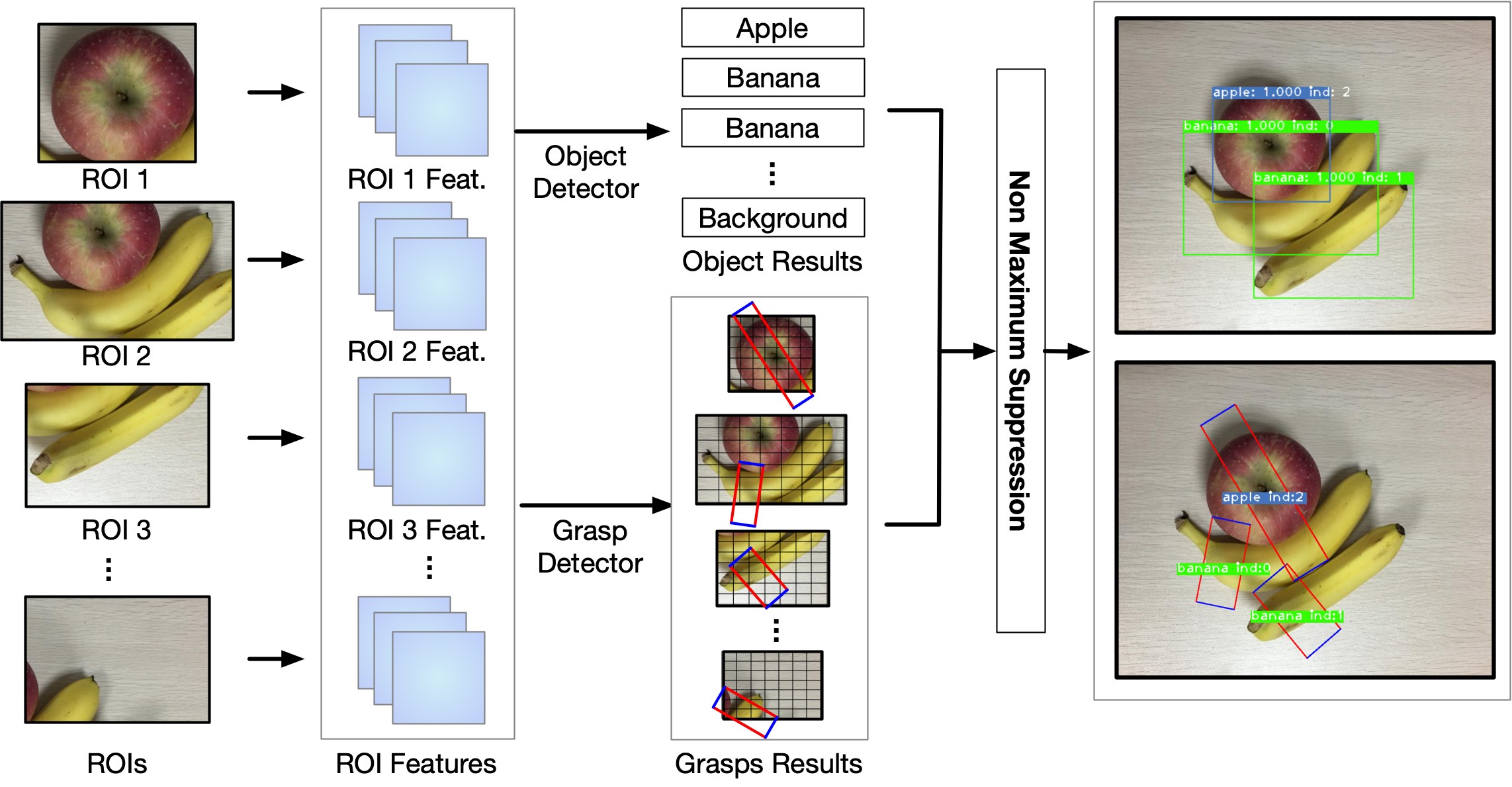}}        
 \caption{ROI-GD with object detector for simultaneous object and grasp detection with the affiliation between them.}
 \label{simulobjgr}
 \end{figure}

In multi-object scenes, in order to construct the affiliation between specified objects and their grasps, objects in the input image need to be detected. Therefore, a new network is designed based on ROI-GD for simultaneous object and grasp detection with the affiliation between them by adding an object detector after ROI pooling layer as shown in Fig. \ref{simulobjgr}.

The object detector and grasp detector both take ROI features as input. In this paper, the object detector is designed as a CNN sub-branch, which classifies the ROIs and refines their locations in the same way with regard to Faster-RCNN\cite{fasterrcnn}. For each ROI, The grasp detector output a set of grasp candidates belonging to it. Afterwards, through NMS, results from the ROIs that most possibly contain objects are preserved including category, location and the best grasp for each object. In our implementation, the object detector includes 3 ResBlocks and is initialized with pretrained weights of layers between C4 and C5 of ResNet-101.

With the proposed network, the robot is able to grasp the specified object in multi-object scenes.
 
\subsection{Relationship with Previous Grasp Detection Algorithms}
 
Several previous works apply deep neural networks for grasp detection\cite{lenzgrasp, redmongrasp, resnetgrasp, guo2017hybrid, chu2018real, zhou2018grasp}. These works transfer object detection algorithms such as SSD\cite{ssd} or Faster-RCNN\cite{fasterrcnn} to grasp detection. They take RGB or RGB-D images as input and treat grasps as a specific kind of object. These algorithms try to detect all potential grasps in the input image, regardless of the owner of the grasps. On single-object grasp datasets such as Cornell Grasp Dataset and Jacquard Dataset, they achieve state-of-the-art performance.

Different from these algorithms, the grasp detector of ROI-GD takes features of ROIs as input instead of the features of the whole image. It is more reasonable for the grasp detector to take object features as input because we hope to detect grasps of objects, not the scenes.

Moreover, though previous grasp detection algorithms like \cite{redmongrasp, resnetgrasp, zhou2018grasp} detect grasps on the whole image, from the view of ROI-based grasp detection, the whole image can be seen as the only ``ROI" in these algorithms. Therefore, it is clear to see that ROI-GD can be easily extended to grasp detection in multi-object scenes while maintaining the advantages of existing grasp detection algorithms, which will be demonstrated in our experiments.

\subsection{Loss Function}

Our loss function includes two parts: ROI detection loss $L_{ROI}$ and grasp detection loss $L_{G}$. 

$L_{ROI}$ is used to train RPN, which is the same as \cite{fasterrcnn}:
\begin{equation}
\begin{split}
L_{ROI} = &-\frac{1}{N_{cls}}\sum_{i=1,...,N_{cls}}{log(p^{(i)})}\\
&+\alpha\frac{1}{N_{reg}}\sum_{i=1,...,N_{reg}}{smoothL1(R^{(i)},R^{(i)}_{GT})}
\end{split}
\end{equation}
where $N_{cls}$ is equal to the anchor number of one mini-batch for training RPN, and $N_{reg}$ is equal to the number of positive samples for training the regressor. $p^{(i)}$ is the probability that the corresponding anchor is classified into the correct category. $R^{(i)}$ and $R^{(i)}_{GT}$ are two 4-dimension vectors representing the location of the predicted ROI and the matched ground truth respectively. In our experiment, $\alpha$ is set to 10 to balance the loss of classification and regression.

For the ROI $R$, the grasp detection loss $L_{G}(R)$ includes grasp regression loss and classification loss:
\begin{equation}
\begin{split}
L_{G}(R) = &\frac{1}{N_G}(-\sum_{i\in Positive}{log(p_g^{(i)})}-\sum_{i\in Negative}{log(p_u^{(i)})}\\
&+\beta\sum_{i\in Positive}{smoothL1(g^{(i)},g^{(i)}_{GT})})
\end{split}
\end{equation}
where $N_G$ is the number of grasp anchors on the ROI for training the grasp detector, $p_g^{(i)}$ and $p_u^{(i)}$ are the probabilities of the grasp anchor classified into ``graspable" and ``ungraspable", $g^{(i)}$ and $g^{(i)}_{GT}$ are two 5-dimension vectors representing the predicted grasp and the ground truth respectively. In our experiment, $\beta$ is set to 10 to balance the loss of classification and regression.

Since the network is trained end-to-end in this paper, for one image, the total loss is defined as follow:
\begin{equation}
loss = L_{ROI}+\lambda{\sum_{R\in ROIs}{L_G(R)}}
\label{loss}
\end{equation}

From Eq. \ref{loss}, we can see that the total loss $loss$ is a multi-task loss of object detection and grasp detection. $\lambda$ is used to balance loss from two separate tasks, which is set to $1/5N_{ROI}$ in this paper and $N_{ROI}$ is the number of ROIs used in grasp detection loss.

\section{Experiment}

\subsection{Dataset}
We evaluate the performance of our method in three different grasp datasets: VMRD, Cornell Grasp Dataset, and Jacquard Dataset.

To train and test our algorithm in multi-object scenes, we expand the Visual Manipulation Relationship Dataset (VMRD)\cite{zhang2018visual} with grasps. VMRD originally has 31 categories and 5185 images including more than 17000 object instances and 51000 manipulation relationships. In each image, each object instance has a unique index to be distinguished from the others.

\begin{figure}[t] 
 \center{\includegraphics[width=0.4\textwidth]{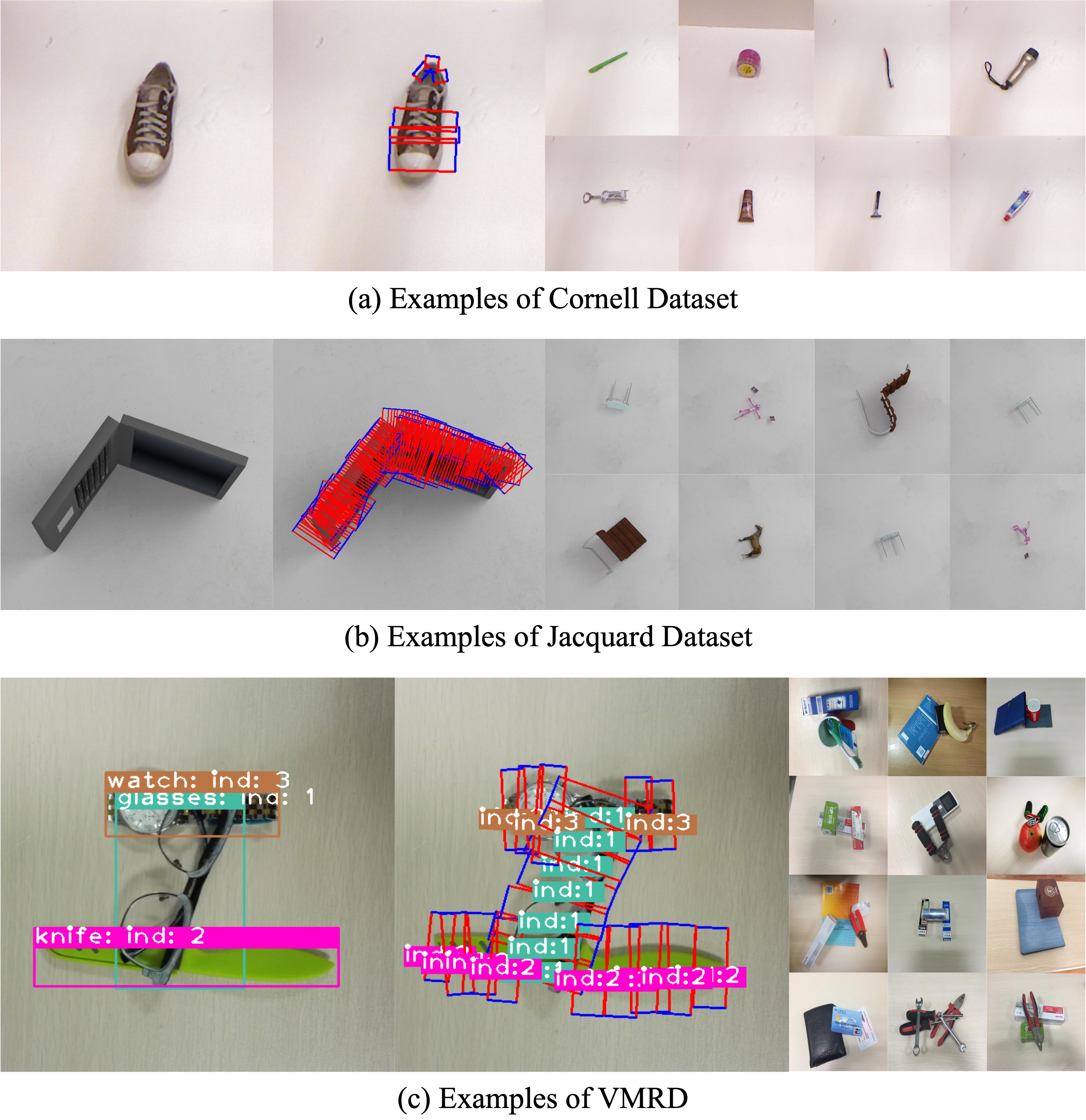}}        
 \caption{Examples of dataset. (a) Examples of Cornell Grasp Dataset; (b) Examples of Jacquard Dataset; (c) Examples of VMRD. There are different sorts of graspable things in VMRD dataset. In each image, each object has a unique index to be distinguished from the others. Grasps belonging to one object are also distinguished by the index from the grasps belonging to the other objects. For example, a labeled image is shown in the left of (c). The knife has an index ``2", hence all grasps with index ``2" in the second image of (c) belong to the knife.}
 \label{dataset}
 \end{figure}

We label 4683 images in VMRD with over $10^5$ grasps in total, much more than grasps in Cornell Grasp Dataset. Some examples are shown in Fig. \ref{dataset} (c). The affiliations between grasps and their owner are also included by using the object index. For example, as shown in the left image of Fig. \ref{dataset} (c), the knife has an index ``2", hence all grasps with index ``2" in the second image belong to the knife. More details can be found at this link\footnote{http://gr.xjtu.edu.cn/web/zeuslan/visual-manipulation-relationship-dataset;jsessionid=E6F2B81979FA7849D9D773A7A389B809}.  The dataset is split into a training set and a testing set. In the training set and the testing set, there are 4233 and 450 images respectively.

For evaluation in single-object scenes, we use Cornell Grasp Dataset and Jacquard Dataset to train and validate our model. Cornell Grasp Dataset includes 885 images and Jacquard Dataset contains 54k images. Some examples of Cornell Grasp Dataset and Jacquard Dataset are shown in Fig. \ref{dataset} (a) and (b). Because there is no object location information in Cornell Grasp Dataset, for training RPN, object images and the corresponding background images are used to generate the location of objects. During experiments, we split the dataset into a training set and a testing set with a ratio of 4:1. For Cornell Grasp Dataset, 5-fold cross validation is used for performance evaluation.

\subsection{Implementation Details}

Our network is implemented on PyTorch. GPUs used to train the network are all GTX1080Ti with 11GB memory. The learning rate of our network is 0.002 and momentum is 0.9. We use a learning rate decay of 0.0002 and 0.000003 for Cornell Grasp Dataset and Jacquard Dataset respectively. The training mini-batch size is set as 4. During training, we use photometric distortion augmentation including random contrast, color space conversion (only RGB and HSV), random saturation and random hue for VMRD and Cornell Grasp Dataset to prevent overfitting. Besides photometric distortion augmentation, we also augment our dataset with horizontal flipped and vertically rotated images (rotate the image by $\pm90^{\circ}$ or $180^{\circ}$). Though the input size is not required to be fixed, we normalize the shorter sides of all images to 600 pixels for VMRD and 800 pixels for Cornell Grasp Dataset and Jacquard Dataset while keeping the aspect ratio of the image unchanged. The mini-batch size for training RPN is 256 for VMRD and 32 for Cornell Grasp Dataset and Jacquard Dataset. During testing, top-300 ROIs are used for VMRD while top-20 ROIs are used for Cornell Grasp Dataset and Jacquard Dataset due to the difference of object number in the input image.
 
\subsection{Metrics}

To demonstrate the advantages of ROI-GD in multi-object scenes, we train our algorithm on VMRD. For evaluation in multi-object scenes, it is not enough to only evaluate the accuracy of detection result like previous works in single-object scenes. We are inspired by the widespreadly used metric \emph{mAP} in object detection, adding consideration of the detected grasp of each object in \emph{mAP} as following:
 
A detection $(o,g)$ is defined as True Positive when it satisfies the following conditions:
\begin{itemize}
\item Each detection $(o,g)$ includes object $o=(B_o,C_o)$ and its Top-1 grasp $g$ simultaneously, where $B_o$ is the detected location of the object and $C_o$ is the category output by the neural network.
\item The object $o$ is classified correctly ($C_o$  is the same as the ground truth) and the bounding box of the object $B_o$ has an IoU larger than 0.5 with the ground truth.
\item The detected grasp $g$ of the object is correct, which means that the detected Top-1 grasp has a Jaccard Index larger than 0.25 and angle difference less than $30^{\circ}$ with at least one ground truth grasp rectangle belonging to the object.
\end{itemize}

For evaluation in single-object scenes, the best grasp in the input image is chosen from the grasps affiliating to the ROI with the highest confidence score. To compute the accuracy of grasp detection, we take the same metric, ``rectangle metric", using in all previous works like in \cite{redmongrasp}: if the best grasp predicted by our network has a Jaccard Index larger than 0.25 with one ground truth, it will be treated as a correct detection. Jaccard Index between the predicted grasp and the ground truth is computed by Eq. \ref{jaccard}:
\begin{equation}
\label{jaccard}
J(g,g_{GT}) = \frac{g\cap{g_{GT}}}{g\cup{g_{GT}}}
\end{equation}

For Cornell Grasp Dataset, we also evaluate our method in two ways: Image-wise split and Object-wise split.
\begin{itemize}
\item {\bf Image-wise split:} The dataset is split randomly. Each image is treated equally. Image-wise split is used to test the performance on seen things.
\item {\bf Object-wise split:} The dataset is split based on object instances. Objects in the testing set cannot be seen during training. Object-wise split is used to test the generalization ability among different kinds of objects.
\end{itemize}

 \subsection{Validation Results}

\begin{table}[t]
\caption{Validation Results of Different Methods on VMRD}
\label{vmrdresults}
\begin{center}
\begin{tabular}{l|c|c}
\hline
\multirow{2}{*}{{\bf Algorithm}} & {\bf mAP with grasp} & {\bf Speed}\\
& \multicolumn{1}{c|}{(\%, Higher is Better)} & (FPS)\\
\hline
Faster-RCNN\cite{fasterrcnn}+FCGN\cite{zhou2018grasp}&  54.5 & {\bf10.3}\\
\hline
ROI-GD, $12\times12$ Anc., k=4  &  {\bf68.2} & \multirow{4}{*} {9.1} \\
ROI-GD, $12\times12$ Anc., k=6 &  67.4 &  \\
ROI-GD, $24\times24$ Anc., k=4  &  66.3 &   \\
ROI-GD, $24\times24$ Anc., k=6  &  63.4 &\\
\hline
\end{tabular}
\end{center}
\end{table}

 \begin{figure}[t] 
 \center{\includegraphics[scale=0.1]{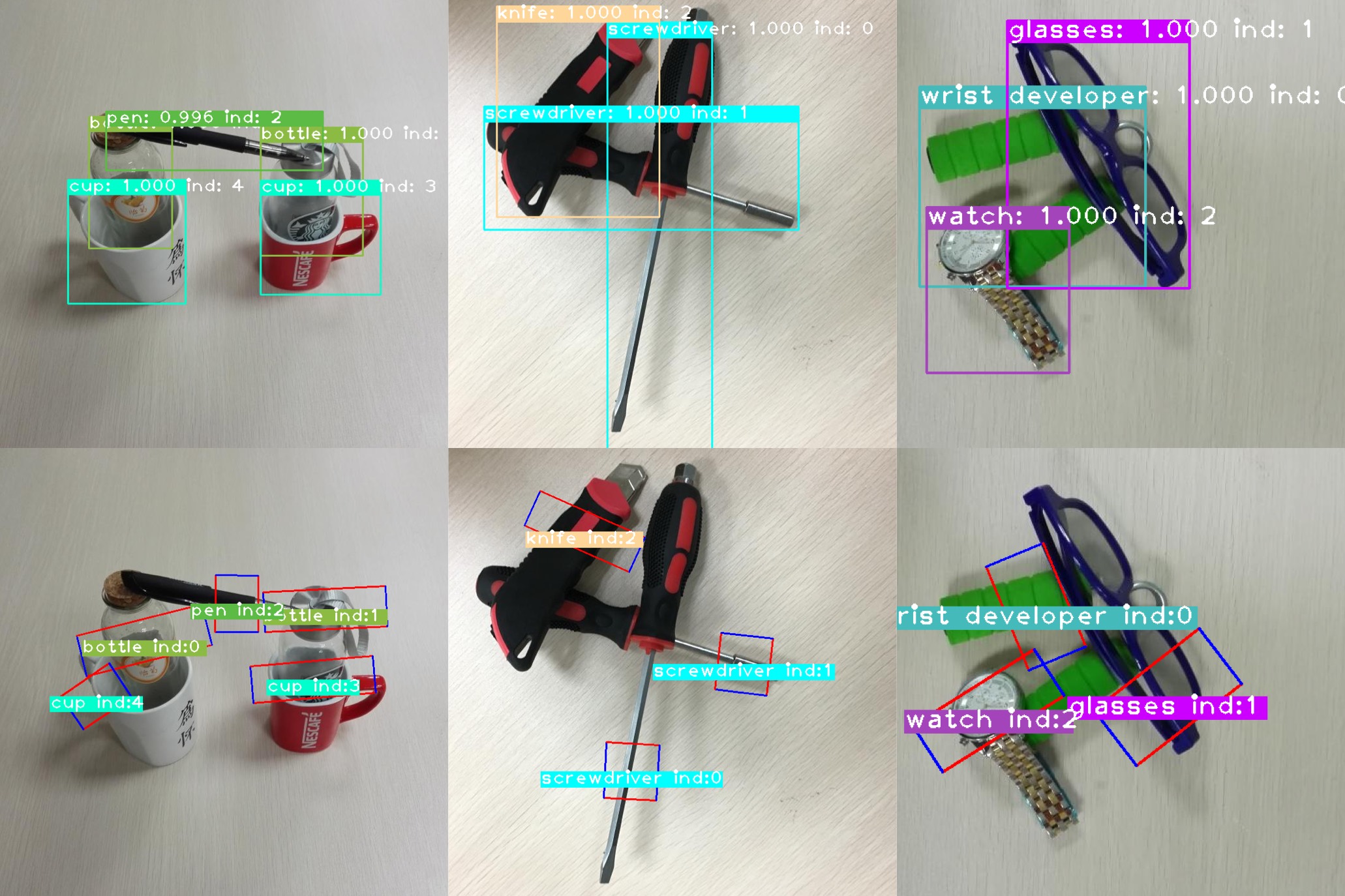}}        
 \caption{Examples of detection results on VMRD. The top row is the object detection results and the second row is the grasp detection results.}
 \label{multiresults}
 \end{figure}
 
{\bf Comparison with Baseline on VMRD} Results on VMRD validation set are shown in Table \ref{vmrdresults} and Fig. \ref{multiresults}. Because it is the first time for simultaneous object and grasp detection on VMRD, the baseline is set as cascading object detector (Faster-RCNN\cite{fasterrcnn}) and grasp detector (FCGN\cite{zhou2018grasp}), which are both state-of-the-art algorithms in object detection and grasp detection respectively. Distance between object center and grasp center is used to assign detected grasps to detected objects. In detail, we choose the closest grasp with confidence score higher than 0.25 to be the grasp of the object, which achieves the highest performance in our testing. 

From Table \ref{vmrdresults}, we can see that ROI-GD improves the performance in multi-object scenes significantly. Lower performance of the baseline is caused by overlaps between objects, especially in the following two situations: (1) grasps are mostly distributed on the edge of the object, such as plates, books, tapes, etc; (2) one object is placed near the center of another object, for example, when a pen is put on a book. Benefiting from ROI-based grasp detection, our algorithm will not suffer from these problems.
 
{\bf Self Comparison} In the experiment, we try different anchor settings to see the influence on our model's performance. Two different anchor sizes are used including $12\times12$ and $24\times24$. From Table \ref{vmrdresults}, we can conclude that the anchor size is an important hyperparameter for network training. Besides, value of $k$ will also influence final results. Larger $k$ will reduce the difference between oriented anchors and matched ground truth grasp rectangles and therefore, reduce the difficulty of regression. However, it will also lead to more hard negatives for graspable classification. Therefore, a proper $k$ is preferred to result in a good performance.

 \begin{table}[t]
\caption{Accuracy of Different Methods on Cornell Grasp Dataset.}
\label{cornellresults}
\begin{center}
\begin{tabular}{l|c|c|c}
\hline
\multirow{2}{*}{\bf{Algorithm}} &  \multicolumn{2}{c|}{\bf{Accuracy*}} &{\bf Speed}\\
\cline{2-3} & \bf{ IW (\%) }&\bf{ OW (\%) }& {\bf  (FPS)} \\
\hline
Fast Search \cite{jiang2011grasp}  & 60.5 & 58.3 & 0.02\\
SAE, struct. reg. Two stage \cite{lenzgrasp}& 73.9 & 75.6 &  0.07\\
MultiGrasp \cite{redmongrasp}&88.0 & 87.1 & 3.31 \\
Multi-model \cite{resnetgrasp} & 89.2 & 89.0 & 16.03 \\
ZF-net \cite{guo2017hybrid} & 93.2 & 89.1 & - \\
STEM-CaRFs \cite{asif2017rgb}  & 88.2 & 87.5 & - \\
GraspNet \cite{asif2018graspnet} & 90.6 & 90.2 & {\bf 41.67}\\
Multiple, M=10 \cite{ghazaei2018dealing} & 91.5 & 90.1 & 17.86 \\
VGG-16 model \cite{chu2018real} & 95.5 & 91.7 &17.24 \\
ResNet-50 model \cite{chu2018real} & 96.0 & 96.1 & 8.33\\
ResNet-50 FCGN \cite{zhou2018grasp} & {\bf 97.7} & 94.9 & 9.89\\
ResNet-101 FCGN \cite{zhou2018grasp} &{\bf 97.7 }& {\bf 96.6} & 8.51\\
\hline
ROI-GD, ResNet-101, RGB   & 93.6 & 93.5 & \multirow{2}{*}{25.16}  \\
ROI-GD, ResNet-101, RGD  & 92.3 & 91.7 &  \\
\hline
\multicolumn{4}{l}{* IW and OW in Accuracy represent Image-wise split and Object-wise}\\
\multicolumn{4}{l}{split respectively.}
\end{tabular}
\end{center}
\end{table}

 \begin{figure}[t] 
 \center{\includegraphics[scale=0.09]{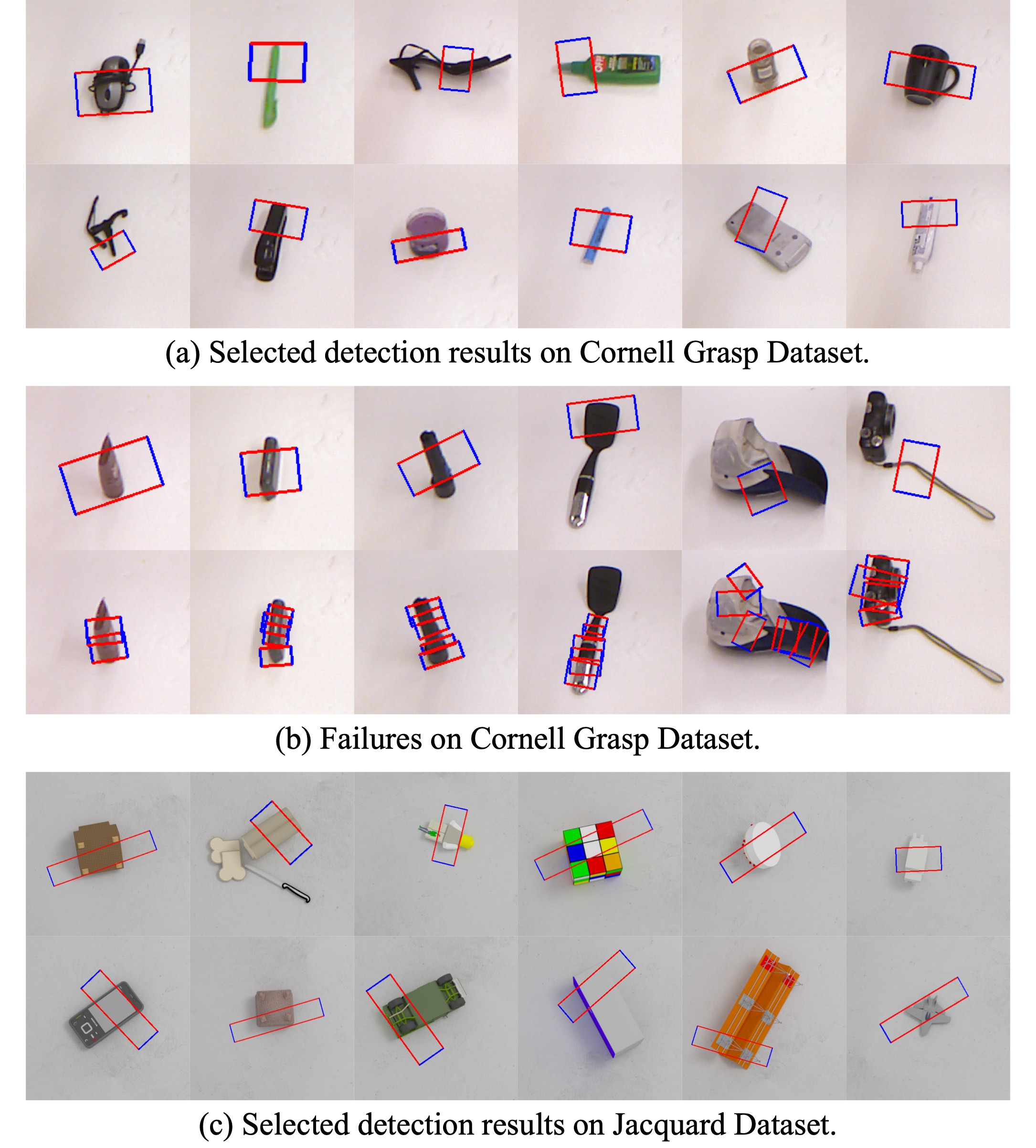}}        
 \caption{Examples of detection results on Cornell Grasp Dataset and Jacquard Dataset. (a) Selected results on Cornell Grasp Dataset. (b) Failures on Cornell Grasp Dataset. The top row is the detection results and the second row is the ground truth. (c) Selected results on Jacquard Dataset.}
 \label{singleresults}
 \end{figure}

\begin{table}[t]
\caption{Accuracy of Different Methods on Jacquard Dataset.}
\label{jacquardresults}
\begin{center}
\begin{tabular}{l|c|c}
\hline
{\bf Algorithm} & {\bf Input} & {\bf Accuracy (\%)} \\
\hline
Jacquard\cite{depierre2018jacquard}& RGB-D &  74.2 \\
\multirow{2}{*}{FCGN, ResNet-101\cite{zhou2018grasp}} & RGB & 91.8\\
& RGD &  92.8\\
\hline
\multirow{2}{*}{ROI-GD, ResNet-101} & RGB & 90.4   \\
& RGD & {\bf 93.6}    \\
\hline
\end{tabular}
\end{center}
\end{table}

{\bf Comparison on Single-object Datasets} Validation results of Cornell Grasp Dataset are demonstrated in Table \ref{cornellresults} and Fig. \ref{singleresults} (a). ROI-GD achieves an accuracy of 93.6\% and 93.5\% in image-wise and object-wise split with a speed of 25.16 FPS. Compared with state-of-the-art algorithm \cite{zhou2018grasp}, there is a performance loss of 4.1\% and 3.1\% in image-wise and object-wise split respectively. 

  \begin{figure*}[!t] 
 \center{\includegraphics[width=\textwidth]{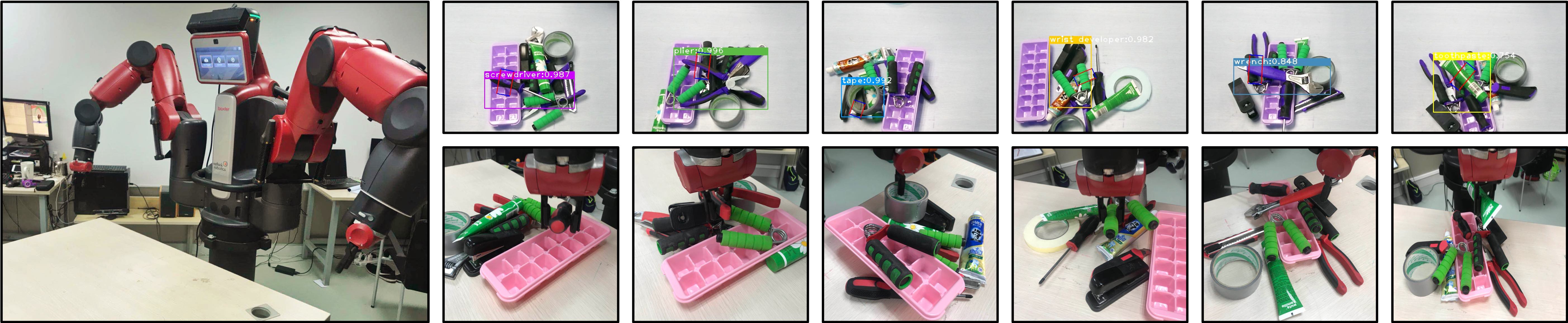}}        
 \caption{Robot experiment. The left picture shows the robot experimental environment. The pictures above are detection results and the pictures below are robotic execution.}
 \label{env}
 \end{figure*}
 
For Jacquard Dataset, the results are listed in Table \ref{jacquardresults}. ROI-GD achieves an accuracy of 93.6\% using RGD images as input. Compared with \cite{depierre2018jacquard}, ROI-GD achieves a 19.4\% gain of the final performance, which is higher than our previous state-of-the-art grasp detection algorithm \cite{zhou2018grasp}. Some examples are shown in Fig. \ref{singleresults} (c).

The difference of performance between Cornell Grasp Dataset and Jacquard Dataset is worth noting. In our visualization as shown in Fig. \ref{singleresults} (b), most incorrect detections on Cornell Grasp Dataset would have been feasible for robotic grasping. In other words, there exist potential grasps unlabeled in this dataset. Therefore, the performance on Cornell Grasp Dataset has bias. Compared with Cornell Grasp Dataset, the bias is lower on Jacquard Dataset due to more grasp labels on each object as shown in Fig. \ref{dataset}.


 
\subsection{Robot Experiment}
 
{\bf Hardware} In our robotic experiment, we use Baxter Robot with two 7-DoF arms designed by RethinkRobotics as the executor. The table is in front of Baxter with its surface and the base of Baxter in the same horizontal plane. The gripper has two parallel fingers with limited active range (about 4cm). Limited by the active range, before each experiment, we set the gripper to a proper position so that the gripper can grasp the target successfully. The camera used in our robotic experiment is Kinect V2, which can collect RGB-D images. The camera is mounted on Baxter head about 70cm higher than the desktop with a depression angle of $75^{\circ}$. Complete environment is shown in Fig. \ref{env}.

{\bf Evaluation Strategy} To evaluate the performance of ROI-GD in real world, we test the success rate of grasping a specific object. The experiments in single-object scenes are similar to the ones in previous works\cite{lenzgrasp, chu2018real}. We put the target alone on the table and make the robot grasp it. In multi-object scenes, different from previous works, we focus on grasping a specific target. Therefore, the target is placed on the table with several other objects as disturbance. To test the generalization ability of our model, we try more complex scenes than VMRD dataset in this part with more cluttered objects. Targets involved in our robotic experiments include apple, banana, wrist developer, tape, toothpaste, wrench, plier and screwdriver. We try 10 times with each of the above objects and record the number of success.

In experiments of multi-object scenes, though targets and their grasps are detected using RGB images, depth information is indispensable to estimate \emph{grasp point} and \emph{approaching vector} from the 5-d grasp configuration in the image coordinate following the method in \cite{lenzgrasp}: the point with minimum depth in a grasp rectangle is chosen as the \emph{grasp point} and the average surface normal near the grasp point is estimated as the \emph{approaching vector}. A limitation of this method is: when grasping an object at the bottom of the pile, it would fail to estimate the \emph{grasp point}, leading to inaccurate grasp height. Therefore, in order to make sure that the target can be successfully grasped, it has to be placed on or next to the other objects instead of being below them.

 \begin{table}[t]
\caption{Robotic Experimental Results of Different Methods}
\label{robresults}
\begin{center}
\begin{tabular}{|l|*{4}{p{1cm}<{\centering}|}}
\hline
{\bf Object} &  \multicolumn{2}{c|}{\bf Prediction} &  \multicolumn{2}{c|}{\bf Grasping} \\
&   \multicolumn{2}{c|}{\bf  Success Rate} &  \multicolumn{2}{c|} {\bf  Success Rate}\\
\cline{2-5}
& Single &Multiple & Single &Multiple\\
\hline
Apple & 10/10 & 10/10& 9/10 & 9/10  \\
Banana & 10/10 &10/10 & 10/10 & 10/10  \\
Wrist Developer & 10/10 & 7/10& 10/10 & 6/10 \\
Tape & 8/10 & 7/10& 8/10 & 7/10   \\
Toothpaste & 10/10 &8/10 & 10/10 & 8/10 \\
Wrench & 10/10 & 10/10& 8/10 & 8/10 \\
Pliers & 10/10 &9/10& 10/10 & 9/10  \\
Screwdriver & 10/10 & 10/10& 9/10 & 10/10  \\
\hline
Total & 97.50\% & 88.75\%& 92.50\% & 83.75\% \\
\hline
\end{tabular}
\end{center}
\end{table}

{\bf Results} Robotic experimental results are shown in Table \ref{robresults} and Fig. \ref{env}. Instead of executing the grasp with the highest graspable score, the grasp closest to the target center with a graspable score more than 0.5 is chosen as the grasp configuration. From Table \ref{robresults}, we can see that in single-object scenes, our algorithm can achieve the overall success rates of 97.5\% and 92.5\% for prediction and execution respectively while in multi-object scenes, the overall success rates are 88.8\% and 83.8\% for prediction and execution respectively on all 8 targets. These results demonstrate that our model can not only deal with grasping in single-object scenes, but also generalize to complex scenes with cluttered objects to grasp the specific target.

\section{Conclusions}

We propose a Region of Interest (ROI) based robotic grasp detection algorithm, named ROI-GD, which detects grasps with consideration of the affiliations between grasps and their owner. It can also be seen as a generalization of the existing grasp detection algorithms. ROI-GD detects grasps on ROI features instead of the features of the whole image. To train the network in multi-object scenes, we contribute a multi-object grasp dataset based on VMRD. Experimental results on Cornell Grasp Dataset, Jacquard Dataset and VMRD demonstrate that compared with state-of-the-art grasp detection algorithms, ROI-GD performs much better in multi-object scenes and at the meantime, remains comparable in single-object scenes. Robotic experimental results show that ROI-GD can achieve the overall success rates of 92.5\% and 83.8\% in grasping tasks of single-object and multi-object scenes respectively.

\addtolength{\textheight}{0cm}   



\section*{ACKNOWLEDGMENT}

This work was supported in part by the key project of Trico-Robot plan of NSFC under grant No.91748208, National Science and Technology Major Project  No. 2018ZX01028-101, key project of Shaanxi province No.2018ZDCXLGY0607, and NSFC No.61573268.

\bibliographystyle{unsrt}
\bibliography{zhbbib}

\end{document}